\documentclass{article}

\usepackage[nonatbib,preprint]{arxiv}
\usepackage{graphicx}
\usepackage{amsmath}
\usepackage{amssymb}
\DeclareMathOperator*{\argmax}{arg\,max}
\usepackage{multirow}
\usepackage[table,xcdraw]{xcolor}

\usepackage[utf8]{inputenc} 
\usepackage[T1]{fontenc}    
\usepackage{hyperref}       
\usepackage{url}            
\usepackage{booktabs}       
\usepackage{amsfonts}       
\usepackage{nicefrac}       
\usepackage{microtype}      
\usepackage{xcolor}         

\title{Transferable Sparse Adversarial Attack}

\author{%
  Ziwen He, Wei Wang\thanks{Corresponding author.}, Jing Dong, Tieniu Tan\\
  Center for Research on Intelligent Perception and Computing, NLPR, CASIA\\
  \texttt{ziwen.he@cripac.ia.ac.cn,\{wwang,jdong,tnt\}@nlpr.ia.ac.cn} \\
}

\begin{document}

\maketitle

\begin{abstract}
  Deep neural networks have shown their vulnerability to adversarial attacks. In this paper, we focus on sparse adversarial attack based on the $\ell_0$ norm constraint, which can succeed by only modifying a few pixels of an image. Despite a high attack success rate, prior sparse attack methods achieve a low transferability under the black-box protocol due to overfitting the target model. Therefore, we introduce a generator architecture to alleviate the overfitting issue and thus efficiently craft transferable sparse adversarial examples. Specifically, the generator decouples the sparse perturbation into amplitude and position components. We carefully design a random quantization operator to optimize these two components jointly in an end-to-end way. The experiment shows that our method has improved the transferability by a large margin under a similar sparsity setting compared with state-of-the-art methods. Moreover, our method achieves superior inference speed, 700$\times$ faster than other optimization-based methods. The code is available at \url{https://github.com/shaguopohuaizhe/TSAA}.
\end{abstract}

\section{Introduction}
\label{sec:1}


In the past few years, deep neural networks (DNNs) have been widely used in many computer vision tasks such as image classification~\cite{Krizhevsky2017}, object detection~\cite{Girshick2014} and action recognition~\cite{Simonyan2014}, etc. However, some studies~\cite{Szegedy2013IntriguingPO, Goodfellow2014ExplainingAH} have shown that DNNs can easily be fooled by adversarial examples which are crafted by maliciously adding designed perturbations to the inputs. These examples are imperceptible to human eyes but result in wrong outputs, posing a potential threat to face recognition~\cite{Dong2019a}, autonomous driving~\cite{Cao2019} and other real-world applications~\cite{Grosse2017,Eykholt2018}.

In most cases, an adversarial disturbance is constrained by the $\ell_p$ norm distance due to its concise mathematical expression. Much prior work~\cite{Szegedy2013IntriguingPO, Goodfellow2014ExplainingAH,RN49,RN53,Madry2017TowardsDL} focuses on adversarial attack under $\ell_\infty$ or $\ell_2$ constraints. Different from these attack methods disturbing all pixels of the image, sparse adversarial attack~\cite{dong2020greedyfool,modas2019sparsefool,croce2019sparse} only perturbs a few pixels but possibly with large modifications. The perturbations are indeed visible but do not alter the semantic content, and can even be applied in the physical world (adversarial patch)~\cite{brown2017adversarial}. For comprehensively assessing the model robustness, it is equally important to explore this category of attack. Among them, $\ell_0$ based attack attracts more attention. It is a typical NP-hard problem. To address this problem, researchers have many attempts under both white-box and black-box settings.

However, existing $\ell_0$ based methods~\cite{papernot2016limitations,croce2019sparse,modas2019sparsefool,dong2020greedyfool} still suffer from such a major issue: they have low transferability~\cite{Papernot2016TransferabilityIM,Papernot2016PracticalBA,Liu2016DelvingIT} when performing attack on black-box models. The transferability means that adversarial examples crafted on one model can successfully attack another model with different architecture and parameters. 
It plays an important role in black-box adversarial attacks. To fool a black-box model, attackers use a substitute model to craft adversarial examples and feed them into the target model. While $\ell_2$ and $\ell_\infty$ norm constrained adversarial examples~\cite{RN53,Liu2016DelvingIT,Jiadong2020,Dong2019} can achieve high transferability across different architectures, it is still unknown how transferable the $\ell_0$ sparse adversarial examples can be. 

In this paper, we explore this question and propose a method to generate transferable sparse adversarial examples. Previous $\ell_0$ based attack methods rely on the target model's accurate gradient information or its approximation, causing the generated adversarial examples overfitting the target model. To remedy this problem, we introduce a trainable generator architecture. It learns to translate a benign image into an adversarial example. Benefited from a large number of data in the training stage, the adversarial image does not overfit the fixed white-box model but can fool many black-box models. 

Previous generator-based methods~\cite{BalujaF17,PoursaeedKGB18,XiaoLZHLS18,SongSKE18,MopuriUB18,Naseer0KKP19} can only craft dense perturbations. To utilize a generator to craft sparse adversarial examples, we elaborately design a framework. In the proposed framework, the origin image is fed into the generator and the output adversarial image is fast obtained through only one feedforward inference without gradient backpropagation. The significant difference from previous generator-based methods is the sparsity of perturbations. Our framework decouples the adversarial perturbation into two components which control distortion magnitude and perturbed pixel location respectively. Then we apply sparse regularization on the location map to achieve a satisfactory sparsity. Each point of the map is a binary value, i.e., 0 or 1, causing a discrete optimization problem. To optimize sparse perturbations in the training stage by an end-to-end way, we design a special 0-1 quantization operator which also ensures the consistency of training and test stages. 

Experiments on the ImageNet~\cite{deng2009imagenet} dataset show that the transferability of our method is better than state-of-the-art methods. 
For images in the ImageNet, when there is no $\ell_\infty$ norm constraint, the recently superior GreedyFool~\cite{dong2020greedyfool} needs to perturb 0.67$\%$ pixels to get 15.09$\%$ fooling rate for transferring the InceptionV3 (IncV3)~\cite{Szegedy2016} to Resnet50 (Res50)~\cite{he2016deep}. In contrast, our method only needs to perturb 0.46$\%$ pixels to achieve the 63.76$\%$ fooling rate.
When there is a constraint $\ell_\infty=10$, our method only needs to modify 14.47$\%$ pixels to achieve a 45.32$\%$ transferability from IncV3 to Res50, while GreedyFool needs 18.19$\%$ pixels to obtain only 10.67$\%$ transferability. Meanwhile, our method demonstrates a much faster inference speed than state-of-the-art methods. To get a similar transferability from Res50 to IncV3 under the same sparsity setting, our method only needs 6 milliseconds to craft an adversarial image on average, while GreedyFool needs 20.49 seconds.

To summarize, the main contributions of this paper are twofold: 1) We propose an $\ell_0$ based sparse adversarial attack framework for better transferability. 
2) We conduct extensive experiments to evaluate the transferability of sparse adversarial attacks. Results on Imagenet demonstrate the superior performance of our method, including transferability and speed. For both with or without $\ell_\infty$ norm constraint, our method outperforms state-of-the-art methods by a large margin.

\section{Related work}
\label{gen_inst}

\paragraph{White-box sparse attack.}
For white-box attack, JSMA~\cite{papernot2016limitations} proposes to select the most effective pixels on the adversarial saliency map, which is used to identify the impact of features on output classification. PGD$_0$~\cite{croce2019sparse} proposes to project the adversarial noise generated by the well-known PGD~\cite{Madry2017TowardsDL} to the $\ell_0$-ball. SparseFool~\cite{modas2019sparsefool} converts the $\ell_0$ constraint problem into an $\ell_1$ constraint problem and exploits the boundaries’ low mean curvature to compute adversarial perturbations. ADMM$_0$~\cite{zhao2018admm} utilizes the alternating direction method of multipliers method~\cite{wu2018ell} to separate the $\ell_0$ norm and the adversarial loss and facilitate the optimization of the sparse attack. SAPF~\cite{FanWLZLLY20} formulates the sparse attack problem as a mixed integer programming to jointly optimize the binary selection factors and continuous perturbation magnitudes of all pixels, with a cardinality constraint on selection factors to explicitly control the degree of sparsity. GreedyFool~\cite{dong2020greedyfool} builds on the greedy algorithm and introduces a GAN-based distortion map for better invisibility. 
These methods contribute to achieve a high white-box attack success rate under a low sparsity setting. We take a further deep step to explore the transferability obtained by performing attack on substitute models.

\paragraph{Black-box sparse attack.}
When it comes to black-box attack, One Pixel Attack~\cite{su2019one} and Pointwise Attack~\cite{SchottRBB19} propose to apply evolutionary algorithms to achieve extremely sparse perturbations. 
CornerSearch~\cite{croce2019sparse} proposes to select the most effective subset of pixels by testing the score of target models after changing one pixel's value to one of the 8 corners of the RGB color cube.
Recently, Sparse-RS~\cite{croce2020sparse} proposes a framework based on random search for score-based sparse attacks in the black-box setting.
GeoDA~\cite{rahmati2020geoda} presents a geometric framework based on the observation that the decision boundary of deep networks usually has a small mean curvature near the data samples and achieves the best fooling rate with a limited query budget. These methods need to query the target model for several times, while our method using local substitute model to craft adversarial examples without any query.

\paragraph{Generator-based attack.}
Some work~\cite{BalujaF17,PoursaeedKGB18,XiaoLZHLS18,SongSKE18,MopuriUB18,Naseer0KKP19} adopts an image-to-image generator architecture in order to learn a mapping from the input image to a perturbed output image such that the perturbed image cannot be distinguished from the benign image for a classification model. 
ATN~\cite{BalujaF17} and GAP~\cite{PoursaeedKGB18} use trainable deep neural networks for transforming images to adversarial examples and perturbations respectively. AdvGAN~\cite{XiaoLZHLS18} applies generative adversarial networks to craft visually realistic perturbations. 
Song et al.~\cite{SongSKE18} synthesize unrestricted adversarial examples entirely from scratch by training a conditional generator. 
Mopuri et al.~\cite{MopuriUB18} present a 
generative model that utilizes the acquired class impressions to learn crafting universal adversarial perturbations. Naseer et al.~\cite{Naseer0KKP19} propose a generative framework that learns to generate strong adversaries using a relativistic discriminator. Above methods focus on all-pixels adversarial images. Instead, we propose a generator framework to craft sparse adversarial examples.

\section{Transferable sparse adversarial attack}
\label{headings}

\subsection{Problem analysis}
\label{sec:pa}

Denote $\mathbf{x}$ as one benign image and $y$ as its corresponding ground-truth label. Let $f$ be the target model and thus we have 
$\argmax_c f(\mathbf{x})_c = y$, where $f(\mathbf{x})_c$ is the output logit value for class $c$. To generate an adversarial sample, an adversary adds elaborately designed noise $\boldsymbol{\delta}$ to the original image $\mathbf{x}$. The resultant adversarial image $\mathbf{x}_{adv} = \mathbf{x} + \boldsymbol{\delta}$ is expected to 
satisfy $\argmax_c f(\mathbf{x}_{adv})_c \neq y$. Meanwhile, the adversarial noise $\boldsymbol{\delta}$ should be small enough to guarantee the imperceptibility. In this paper, the constraint on $\boldsymbol{\delta}$ is measured by $\ell_0$ and $\ell_\infty$ norm:
\begin{equation}\label{eq:1}
\begin{aligned}
 & \min_{\boldsymbol{\delta}} \|\boldsymbol{\delta}\|_0 \\
\textit{s.t.} \quad & \argmax_c f(\mathbf{x}+\boldsymbol{\delta})_c \neq y \\
& \|\boldsymbol{\delta}\|_{\infty}<\epsilon 
\end{aligned}.
\end{equation}

where $\epsilon$ is a hyper-parameter to control $\ell_\infty$ norm constraint of adversarial perturbations.
The above setting is non-targeted attack, and for targeted attack the condition is $\argmax_c F(\mathbf{x}_{adv})_c = y_t$, where $y_t$ is the target label. 

In transfer-based attack, attackers use the white-box model $f$ to craft an adversarial example $\mathbf{x}_{adv}$ and feed $\mathbf{x}_{adv}$ into an unknown target model. For previous sparse attack methods, they craft the adversarial example $\mathbf{x}_{adv}$ heavily relying on the gradient information of $f$ with respect to the single image $x$, resulting in an overfitting issue. To alleviate the overfitting, we introduce a generator-based method. The generator learns a mapping between natural images and sparse adversarial images. Thus the generator's parameter is optimized by a data distribution rather than a single image. The increase of training data amount can reduce overfitting and therefore boost transferability. Under this assumption, the question is then transformed into how to design a generator-based framework to solve the problem \ref{eq:1}.

The minimum optimization problem \ref{eq:1} is NP-hard. To solve this problem, many prior work solve the approximate $\ell_1$ constraint problem. For instance, SparseFool~\cite{modas2019sparsefool} is an algorithm that exploits such a relaxation, by adopting an iterative procedure that includes a linearization of the classifier at each iteration, in order to estimate the minimal adversarial perturbation. However, this iterative solution includes non-differentiable steps, thus cannot be used in an end-to-end training on a generator. To solve the approximate $\ell_1$ constraint problem within a gererator-based framework, one idea is to directly add an $\ell_1$ regularization $\boldsymbol{\delta}$. However, the $\ell_1$ regularization on $\boldsymbol{\delta}$ will make the value of perturbation converge around 0 and thus result in a dense solution. If binary quantization is applied on this solution to obtain a real sparse perturbation, the obtained example is probably not adversarial due to information loss. We solve this problem by factorizing the perturbation to the element-wise product of two variables, including a vector which controls perturbation magnitudes and a binary mask which controls where to perturb. Formally,
\begin{equation}
    \boldsymbol{\delta} = \mathbf{r} \odot \mathbf{m},
\end{equation}
where $\mathbf{r}\in \mathbb{R}^N$ denotes the vector of perturbation magnitudes and $\mathbf{m}\in\{0,1\}^N$ denotes the vector of perturbed positions, with $N$ being the data dimension, and $\odot$ represents element-wise product. Then we jointly optimize $\mathbf{m}$ and $\mathbf{r}$, and only apply $\ell_1$ regularization on $\mathbf{m}$ instead of $\boldsymbol{\delta}$. The $\ell_1$ regularization on $\mathbf{m}$ can lead to a sparse map and meanwhile does not affect the optimizaiton of perturbation magnitude for a successful attack. 

In our framework, we utilize two different branches to optimize the two variables respectively. One branch outputs $\mathbf{r}$, which is bounded by a predefined $\ell_\infty$ norm $\epsilon$. The other branch 
outputs the location map $\mathbf{m}$, each element of which is a binary value (0 or 1). One pixel $(i,j)$ is perturbed if $\mathbf{m}_{(i,j)}=1$, otherwise not perturbed. The binary value of $\mathbf{m}$ causes a challenging discrete optimization problem, as it cannot be directly optimized using any gradient-based continuous solver. To solve this problem, we design a 0-1 random quantization operator which translates continuous vector into discrete vector and enables gradient backpropagation. We describe the quantization operator in the next subsection.
\subsection{Framework}

\begin{figure}
  \centering
  \includegraphics[width=\linewidth]{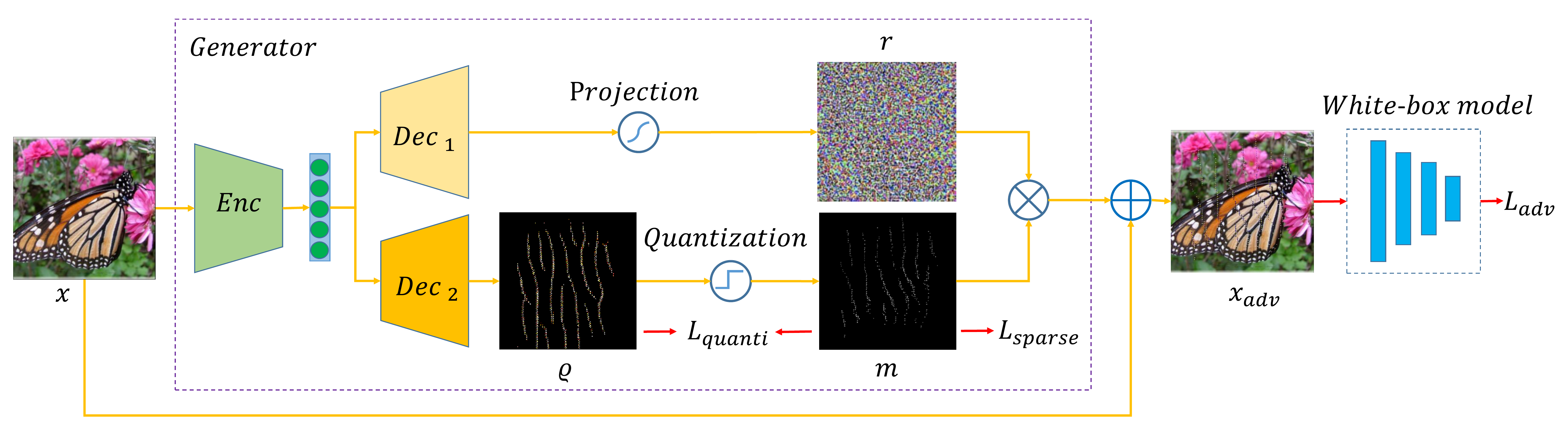}
  \caption{Our framework for generating transferable sparse adversarial examples.}
  \label{fig:1}
\end{figure}

Figure \ref{fig:1} illustrates the overall architecture. A generator is designed to translate a benign image into an adversarial image. Denote the generator as $\mathcal{G}$, the adversarial image is crafted by $\mathbf{x}_{adv} = \mathbf{x} + \mathcal{G}(\mathbf{x})$. The generator mainly includes one encoder and two decoder branches. Firstly the encoder $\mathcal{E}$ takes the original instance $\mathbf{x}$ as its input and generates a latent code $\mathbf{z}=\mathcal{E}(\mathbf{x})$. Then $\mathbf{z}$ is fed into two decoders, denoted as $\mathcal{D}_1$ and $\mathcal{D}_2$.

$\mathcal{D}_1$ is similar to the decoder in $\ell_\infty$ norm generators~\cite{PoursaeedKGB18,Naseer0KKP19}. It outputs a vector which represents perturbation magnitudes. A projection operator is used to bound the vector in a valid range, such as [-255,255]. 
We use a scale operation to achieve the projection and the whole process can be formulated as $\mathbf{r}=\epsilon\cdot \mathcal{D}_1(\mathbf{z})$, where $\epsilon$ is the hyper-parameter to control $\ell_\infty$ norm constraint of adversarial perturbations.

$\mathcal{D}_2$ outputs a vector $\boldsymbol{\varrho}\in[0,1]^N$. To get the discrete vector $\mathbf{m}$ which controls the perturbed pixels' positions, we need to pass $\boldsymbol{\varrho}$ through a binary quantization operator $q$. Normally, a hard label quantization operator is
\begin{equation}
\label{eq:3}
q(\varrho_{i,j})=\left\{\begin{array}{ll}
0 & \varrho_{i, j} \leqslant \tau\\
1 & \varrho_{i, j} > \tau,
\end{array}\right.
\end{equation}
where $\tau$ is a predefined threshold and $\varrho_{i,j}$ represents the value of $\boldsymbol{\varrho}$ at pixel ($i,j$). Obviously, such a quantization operator will cause gradient vanishing if used in training (the derivative of all differentiable points is 0). Therefore, we design an operator including randomness in the training stage. Denote whether to quantize $\varrho_{i,j}$ as a random variable $X\in\{0,1\}$. If $X=1$, $\varrho_{i,j}$ is quantized to 0 or 1 using Equation~\ref{eq:3}, otherwise keep its original value. We make $X$ subject to Bernoulli distribution. The probability distribution is 
\begin{equation}
\label{eq:4}
P(X=k)=p^k(1-p)^{1-k},k\in\{0,1\},
\end{equation}
where $p$ is the probability of $X=1$.
We block gradient backpropagation when $X=1$ and only permit gradient flow when $X=0$. In other words, if one pixel $m_{i,j}$ is not quantized in the training stage, its gradient cannot be backpropageted to $\varrho_{i,j}$. By this way, we make $\varrho_{i,j}$ gradually approaching 0 or 1 and ensure accurate gradient information in the training process. In the inference stage, we set $p$ = 1 and thus all points of $\boldsymbol{\varrho}$ are quantized to a binary value.

Finally, the adversarial image can be obtained by $\mathbf{x_{adv}}=\mathbf{x}+\mathbf{r}\cdot \mathbf{m}$. Once $\mathcal{G}$ is trained on the training data and the white-box model, it can produce perturbations for any input instance to perform a transfer attack.

\subsection{Loss functions}

\paragraph{Adversarial loss.}
In order to achieve our goal of fooling a target black-box model, we need a local substitute model $f$ to supervise the training of generator $\mathcal{G}$. For non-targeted attack, the goal is to generate an adversarial image that is not classified as the ground-truth label. At each training iteration, the generator tries to maximize the adversarial image's probability of a wrong label via a loss function. To achieve this goal, we use the loss function from C$\&$W~\cite{RN49} as the adversarial loss:
\begin{equation}
\label{eq:cw}
    \mathcal{L}_{adv}(\mathbf{x}_{adv}, y, f)=\max \left(f(\mathbf{x}_{adv})_{y}-\max _{i \neq y}\left\{f(\mathbf{x}_{adv})_{i}\right\},-\kappa\right),
\end{equation}
where $\kappa$ is a confidence factor to control the attack strength.
Targeted attack can be achieved simply by replacing the first term with $\max _{i \neq y_t}\left\{f(\mathbf{x}_{adv})_{i}\right\}-f(\mathbf{x}_{adv})_{y_t}$, where $y_t$ is the target label.

\paragraph{Sparse loss.}
For general adversarial attack, perturbation should also
be minimal such that the adversarial image appears as a legitimate image similar to the input image. Here for $\ell_0$ sparse adversarial attack, the constraint on perturbation is a small $\ell_0$ norm. In our framework the perturbation is generated by $\boldsymbol{\delta}=\mathcal{G}(\mathbf{x})=\mathbf{r}\cdot\mathbf{m}$ and $\mathbf{r}$ is continuous, so the degree of sparsity is mainly controlled by $\mathbf{m}$. Since the $\ell_0$ norm is non-differentiable, we cannot incorporate it directly as a regularization term in the objective function. As we discussed in subsection \ref{sec:pa}, the minimization of $\ell_0$ can be relaxed to $\ell_1$. 
Thus we use the $\ell_1$ regularization on $\mathbf{m}$,
\begin{equation}
\label{eq:sparse}
    \mathcal{L}_{sparse}(\mathbf{m})=\|\mathbf{m}\|_1.
\end{equation}
The above function urges perturbation to be as sparse as possible and the resultant perturbation has a dynamic pixel number according to the convergence degree of generator $\mathcal{G}$, hyper-parameters setting, etc.

\paragraph{Quantization loss.}

We have designed a special quantization operator which is in different state in training and test. This may bring some test error due to the information loss caused by quantization. To reduce the gap between training and test, a straightforward solution is to reduce the quantization error of $\boldsymbol{\varrho}$. The quantized parameter should approximate the full-precision parameter as closely as possible, expecting the test performance will be close to that of training.
The quantization loss is:
\begin{equation}
    \mathcal{L}_{quanti}(\boldsymbol{\varrho})=\|\boldsymbol{\varrho}-\mathbf{m}\|_2.
\end{equation}

The overall loss is:
\begin{equation}
\label{eq:10}
    \mathcal{L}=\mathcal{L}_{adv}+\lambda_s\mathcal{L}_{sparse}+\lambda_q\mathcal{L}_{quanti},
\end{equation}
where $\lambda_s$ and $\lambda_q$ are hyper-parameters controlling the relative importance of sparse and quantization losses, respectively. $\mathcal{L}_{sparse}$ and $\mathcal{L}_{quanti}$ encourage the generated perturbation to be sparse in the inference stage, while $\mathcal{L}_{adv}$ optimizes for a high attack success rate. After training, the generator $\mathcal{G}$ can generate an adversarial image for any input image and can
be tested on any black-box model.

\section{Experiments}
\label{others}

\paragraph{Experimental setup.}

We perform experiments on the widely used ImageNet~\cite{deng2009imagenet} dataset. 
We generate adversarial samples by attacking an Inception-v3 (IncV3) model~\cite{Szegedy2016} and a Resnet50 (Res50) model~\cite{he2016deep} respectively. 
For our proposed method, we use the source model to supervise the training of generator and directly translate original image into adversarial image during inference time. When using IncV3 as the source model, the input image size is cropped to 299$\times$299 while for Res50 is 224$\times$224. 
For target models, we also use a VGG16 without batch normalization (VGG16)~\cite{simonyan2014very} and a Densenet161 (Dense161)~\cite{huang2017densely}. 
To make an accurate comparison, we generate adversarial samples with 5000 images randomly selected from the ImageNet validation set. 
The natural accuracy of all models on the test set is higher than 90$\%$.

\paragraph{Implementation of generator.}
In our experiments we use a popular residual network generator architecture to translate natural images into adversarial examples. Similar networks are often used in an image-to-image translation task~\cite{isola2017image,johnson2016perceptual,zhu2017unpaired}. Since the residual network generator is fully convolutional, it can be applied to images of any resolution. The input and output are both color images of identical shape. For the encoder, it contains three stride-2 convolutions and six residual blocks~\cite{he2016deep}. 
For both two decoders, we use the same architecture which consists of three 1/2-strided convolutions, except that the channel of output layer is 3 for Dec$_1$ and 1 for Dec$_2$ respectively. The generator in the following experiments is trained with the whole ImageNet training set.

\paragraph{Baselines.}
Throughout our experiments we rely on three standard sparse attack strategies which can be adapted to transfer-based setting: PGD$_0$~\cite{croce2019sparse}, SparseFool~\cite{modas2019sparsefool}, and GreedyFool~\cite{dong2020greedyfool}. We use the official implementation of these methods. PGD$_0$ needs a pre-defined number of perturbation pixels and calculate the fooling rate under such a pre-defined number, while SparseFool and GreedyFool perturb an image with a dynamic pixel number and run until a successful attack or iteration upper bound. Thus for a fair comparison, we report results after finetuning hyper-parameters to get a comparable sparsity. 

\paragraph{Hyper-parameters.}
For PGD$_0$, it does not find an adversarial example for each test point due to the fixed maximum number of pixels that can be modified. Therefore, we directly pre-define a sparsity number similar to our method for it.
For SparseFool, $\lambda$ is its only hyper-parameter and can be easily adjusted to meet the corresponding needs in terms of fooling rate, sparsity, and complexity. For GreedyFool, since it also uses the C$\&$W loss, the $\kappa$ in Equation \ref{eq:cw} controls the attack strength. For example, when $\kappa$ = 0, the attack stops once the generated adversarial sample is adversary. When $\kappa >$ 0, pixels keep increasing until the logit difference $\max _{i \neq y}\left\{f(\mathbf{x}_{adv})_{i}\right\}-f(\mathbf{x}_{adv})_{y}>\kappa$. Since $\kappa$ is contradiction with the reduce stage of GreedyFool, we follow the setting in GreedyFool to only use the increasing stage of GreedyFool. 
We finetune $\lambda$ and $\kappa$ until their sparsity is similar to our method and then use the fooling rate on black-box target model for evaluation. For the sake of completeness, we also report results which inherit all hyper-parameter settings from their respective papers.
For our method, we set $\tau$ = 0.5 in Equation \ref{eq:3}, $p$ = 0.5 in Equation \ref{eq:4}, $\kappa$ = 0 in C$\&$W loss. $\lambda_s$ and $\lambda_q$ in Equation \ref{eq:10} are finetuned in different sparsity setting.

\subsection{Transferability evaluation}
\label{sec:4.1}
In this section, we evaluate the transferability under different perturbation $\ell_\infty$ norm constraint. We report both fooling rates on white-box and black-box models under a similar sparsity setting. A higher fooling rate on black-box target models indicates a better adversary. The sparsity is the average proportion of disturbed pixels in a single image. We also report the average speed of computing an adversarial image.

\paragraph{Non-targeted result.}
Non-targeted attack results for $\ell_\infty=255$ and $\ell_\infty=10$ on ImageNet are shown in Table \ref{table:1} and Table \ref{table:2}, respectively.
For $\ell_\infty=255$ setting, under a similar sparsity, the inference speed of our method is only 0.006 seconds to achieve attack, which is several orders of magnitude faster than other methods. As the sparsity increases, the transferability of baselines increases, while our method is still better than others with a large margin. 
When the source model is IncV3, comparing PGD$_0$, SparseFool ($\lambda$ = 10) and GreedyFool ($\kappa$ = 15) with ours, our method obtains better transferability with lower sparsity and faster inference speed. When it comes to Res50 as the source model, we find that our method still outperforms other methods in most settings. The only exception is when transferring to IncV3 and the reason may be the difference of image size during training. 
When $\ell_\infty$ = 10, the inference process of our method is still very fast. The transferability of our method is also still better than others with a large margin except when transferring from Res50 to IncV3. 


\begin{table}
\caption{$\ell_\infty=255$ constrained non-targeted attack transferability comparison on ImageNet dataset. The best speed and transfer rate are shown in bold. `*' means white-box setting.}
\label{table:1}
\centering
\begin{tabular}{llcccccc}
\toprule
&    &    &     &     &      &    &   \\
\multirow{-2}{*}{Source} & \multirow{-2}{*}{Method}                                    & \multirow{-2}{*}{\begin{tabular}[c]{@{}c@{}}Sparsity\\ (\%)\end{tabular}} & \multirow{-2}{*}{\begin{tabular}[c]{@{}c@{}}Speed\\ (s)\end{tabular}} & \multirow{-2}{*}{\begin{tabular}[c]{@{}c@{}}IncV3\\ (\%)\end{tabular}} & \multirow{-2}{*}{\begin{tabular}[c]{@{}c@{}}Res50\\ (\%)\end{tabular}} & \multirow{-2}{*}{\begin{tabular}[c]{@{}c@{}}VGG16\\ (\%)\end{tabular}} & \multirow{-2}{*}{\begin{tabular}[c]{@{}c@{}}Dense161\\ (\%)\end{tabular}} \\ \midrule
& PGD$_0$  & { 0.56}  & {62.19}   & {56.50*}   & {21.95}  & { 23.60} & {9.69}  \\ 
 & SparseFool  & { 0.26}  & {13.80}   & { 99.90*}   & { 7.34}  & {14.24} & {5.04}  \\ 
 & SparseFool($\lambda$=10)  & {0.52}  & {8.80}   & {100.00*}   & {11.76}  & {24.50} & {6.96}  \\ 
  & GreedyFool  & { 0.11} & {7.05} & { 100.00*} & {2.16}  & {5.38} & {1.38}  \\ &
  \begin{tabular}[c]{@{}l@{}}GreedyFool($\kappa$=15)\end{tabular} & { 0.67}  & {63.12}   & { 100.00*} & {15.09} & { 26.37} & {11.94} \\
\multirow{-6}{*}{IncV3}  & Ours  & { 0.46}   & {\textbf{0.006}}  & { 61.24*} & { \textbf{63.76}}  & { \textbf{85.94}}  & { \textbf{46.22}}  \\\midrule
& PGD$_0$  & { 0.60}  & {18.62}   & {20.54}   & {75.74*}  & { 43.50} & {16.72}  \\ 
  & SparseFool  & { 0.41} & {14.23}  & {21.56} & {98.74*}  & {25.34}  & {9.90} \\ 
   & SparseFool($\lambda$=10)  & { 0.66} & {5.81}  & {27.18} & {100.00*}  & {35.40}  & {13.56} \\ 
& GreedyFool & { 0.22} & {6.74} & {2.52} & {100.00*}  & {8.88}  & {1.80}\\ 
& \begin{tabular}[c]{@{}c@{}}GreedyFool($\kappa$=15)\end{tabular} & { 0.75}  & {22.87}  & {\textbf{29.12}}  & {100.00*}  & { 43.88} & {30.09} \\ 
\multirow{-6}{*}{Res50}  & Ours  & { 0.59} & { \textbf{0.006}}  & {25.90}  & {79.04*}  & { \textbf{85.96}}   & { \textbf{60.18}}  \\
\bottomrule
\end{tabular}
\end{table}

\begin{table}
\caption{$\ell_\infty=10$ constrained non-targeted attack transferability comparison on ImageNet dataset. The best speed and transfer rate are shown in bold. `*' means white-box setting.}
\label{table:2}
\centering
\begin{tabular}{llcccccc}
\toprule
&    &    &     &     &      &    &   \\
\multirow{-2}{*}{Source} & \multirow{-2}{*}{Method}                                    & \multirow{-2}{*}{\begin{tabular}[c]{@{}c@{}}Sparsity\\ (\%)\end{tabular}} & \multirow{-2}{*}{\begin{tabular}[c]{@{}c@{}}Speed\\ (s)\end{tabular}} & \multirow{-2}{*}{\begin{tabular}[c]{@{}c@{}}IncV3\\ (\%)\end{tabular}} & \multirow{-2}{*}{\begin{tabular}[c]{@{}c@{}}Res50\\ (\%)\end{tabular}} & \multirow{-2}{*}{\begin{tabular}[c]{@{}c@{}}VGG16\\ (\%)\end{tabular}} & \multirow{-2}{*}{\begin{tabular}[c]{@{}c@{}}Dense161\\ (\%)\end{tabular}} \\ \midrule
& PGD$_0$  & {14.54}  & {63.28}   & { 97.89*}   & {9.70}  & {12.73} & {8.16}  \\ 
& SparseFool  & {1.65}  & {29.60}   & { 99.98*}   & { 4.94}  & {9.10} & {4.08}  \\ 
 & SparseFool($\lambda$=10)  & {12.56}  & {83.85}   & { 100.00*}   & {7.99}  & {12.63} & {11.40}  \\ 
  & GreedyFool  & { 0.55} & {2.79} & { 100.00*} & {0.94}  & {0.58} & {2.08}  \\ 
  & GreedyFool($\kappa$=40) & {18.19}  & {84.67}  & { 100.00*} & {10.67} & {11.24} & {6.67}\\ 
\multirow{-6}{*}{IncV3}  & Ours  & {14.47}   & {\textbf{0.006}}  & { 87.72*} & { \textbf{45.32}}  & { \textbf{50.38}}  & { \textbf{28.98}}  \\\midrule
& PGD$_0$  & {9.96}  & {18.36}   & {11.38}   & {99.54*}  & {21.42} & {20.74}  \\ 
& SparseFool  & {1.27} & {7.80}  & {2.92} & {99.96*}  & { 2.94}  & {2.02}\\
& SparseFool($\lambda$=15)  & {9.72}  & {36.00}   & { 11.87}   & {100.00*}  & {13.39} & {14.23}  \\ 
& GreedyFool  & {0.59} & {1.22}  & {3.20} & {100.00*}  & { 2.76}  & {1.42}\\ 
& \begin{tabular}[c]{@{}c@{}}GreedyFool($\kappa$=30)\end{tabular} & {12.64}  & {20.49}  & {\textbf{ 12.35}}  & {100.00*}  & {17.09} & {20.89} \\ 
\multirow{-6}{*}{Res50}  & Ours  & {10.52} & { \textbf{0.006}}  & {9.20}  & {72.90*}  & { \textbf{39.48}}   & { \textbf{51.18}}  \\
\bottomrule
\end{tabular}
\end{table}

\paragraph{Targeted attack result.}
We then explore the much harder targeted attack. As SparseFool operates as a non-targeted attack, here we compare with PGD$_0$ and GreedyFool on ImageNet dataset in Table \ref{table:7}. Running on all 1000 target classes of ImageNet is too time consuming, so we randomly choose two target classes including `vulture'(ID:23) and `bubble'(ID:971). We find it is hard for all methods to find transferable targeted sparse adversarial samples. Though, our method apparently improves transferability when the target is `bubble'. For different target classes, the transfer rate varies mainly due to the diverse similarity of decision boundaries between two models~\cite{Liu2016DelvingIT}. Analyzing which target classes leading to strong transferability is left for an intersting future work.
\begin{table}
	\caption{Targeted attack transferability comparison. The source model is IncV3 and attacks are performed on ImageNet dataset, with $\ell_\infty=255$ constraint. The best speed and transfer rate are shown in bold. `*' means white-box setting.}
	\label{table:7}
	\centering
	\begin{tabular}{llcccccc}
		\toprule
		&    &    &     &     &      &    &   \\
		\multirow{-2}{*}{\begin{tabular}[c]{@{}c@{}}Target\\ class \end{tabular}} & \multirow{-2}{*}{Method}                                    & \multirow{-2}{*}{\begin{tabular}[c]{@{}c@{}}Sparsity\\ (\%)\end{tabular}} & \multirow{-2}{*}{\begin{tabular}[c]{@{}c@{}}Speed\\ (s)\end{tabular}} & \multirow{-2}{*}{\begin{tabular}[c]{@{}c@{}}IncV3\\ (\%)\end{tabular}} & \multirow{-2}{*}{\begin{tabular}[c]{@{}c@{}}Res50\\ (\%)\end{tabular}} & \multirow{-2}{*}{\begin{tabular}[c]{@{}c@{}}VGG16\\ (\%)\end{tabular}} & \multirow{-2}{*}{\begin{tabular}[c]{@{}c@{}}Dense161\\ (\%)\end{tabular}} \\ \midrule
		& PGD$_0$  & { 0.56}  & {56.60}   & {0.00*}   & {0.00}  & { 0.00} & {0.00}  \\ 
		& GreedyFool  & {0.49} & {35.99} & { 100.00*} & {0.00}  & {0.00} & {0.02}  \\ 
		\multirow{-3}{*}{\begin{tabular}[c]{@{}l@{}}`vulture'\\(ID:23)\end{tabular}}  & Ours  & {0.79}   & {\textbf{0.006}}  & { 34.28*} & {\textbf{0.10}}  & {\textbf{0.08}}  & {\textbf{0.40}}  \\
		\midrule
		& PGD$_0$  & { 0.56}  & {58.53}   & {0.00*}   & {2.25}  & { 6.50} & {0.38}  \\ 
		& GreedyFool & { 0.42} & {25.42} & {99.90*} & {0.10}  & {0.16}  & {0.06}\\  
		\multirow{-3}{*}{\begin{tabular}[c]{@{}l@{}}`bubble'\\(ID:971)\end{tabular}}  & Ours  & {0.55} & { \textbf{0.006}}  & {35.38*}  & {\textbf{10.38}}  & { \textbf{9.08}}   & { \textbf{3.66}}  \\
		\bottomrule
	\end{tabular}
\end{table}

\subsection{Comparison with generator-based methods}
Our framework is based on a generator architecture, which directly outputs a sparse perturbation. Here we compare it with generator-based dense attack methods, including GAP~\cite{PoursaeedKGB18} and cross-domain perturbations~\cite{Naseer0KKP19}. We follow their work to set $\ell_\infty$ = 10 and use IncV3 as the source model. Results are shown in Table \ref{table:6}. Both dense attacks achieve a high transfer rate with near all-pixel perturbations. We tune the $\lambda_s$ in Equation \ref{eq:10} to obtain results with diverse sparsity. With $\lambda_s$ increasing, both the sparsity and transfer rate drops, showing the number of modified pixels affects transferability when the pixel budget is limited. However, when $\lambda_s$ = 5$\times$10$^{-6}$, the transfer rate of our method is competitive to the two dense attacks while our perturbation is sparser. This demonstrates that some modified pixels of dense attacks are redundant for a successful attack. Therefore, sparse adversarial examples as a natural consequence of removing redundancy can be as transferable as dense examples.

\begin{table}
  \caption{Comparison with generator-based dense attacks. Results are sparsity($\%$) and fooling rate($\%$) on different models. `*' means white-box setting.}
  \label{table:6}
  \centering
  \begin{tabular}{lccccc}
    \toprule
    Method & Sparsity & IncV3 & Res50 & VGG16 & Dense161           \\
    \midrule
    GAP  & {98.98}  & {91.62*} & {82.42}  & {86.26}  & {72.14}  \\
    Cross-domain perturbations  & {99.91}  & {98.10*} & { 88.96}  & {95.86}  & {83.76}  \\
    \midrule
    Ours($\lambda_s$=5$\times$10$^{-6}$) & 39.26 & 99.06* & 86.84 & 90.54 & 76.82\\
    Ours($\lambda_s$=1$\times$10$^{-5}$) & 27.54 & 97.82* & 70.90 & 75.82 & 54.34\\
    Ours($\lambda_s$=1$\times$10$^{-4}$)& {14.47} & { 87.72*} & { 45.32}  & { 50.38}  & { 28.98}  \\
    Ours($\lambda_s$=2$\times$10$^{-4}$)& {7.98} & {71.36*} & {32.82}  & {36.00}  & { 20.60}  \\
    \bottomrule
  \end{tabular}
\end{table}

\subsection{Ablation study}

We further analyze the contributions of key parts in our framework toward sparsity and transferability. Here we set $\ell_\infty$ = 255 and use IncV3 as the source model. Results are shown in Table \ref{table:5}. 

\paragraph{Effects of decoupling.}
In the following, we evaluate the effect of modules in our framework. To prove the importance of decoupling, we delete the path of Dec$_2$ in the proposed framework and then use single path of Dec$_1$ to output $\boldsymbol{\delta}$, on which sparse loss $\mathcal{L}_{sparse}(\boldsymbol{\delta})$ is added.  We mark this setting as `w/o decoupling' and its sparsity is 82.88$\%$, demonstrating that directly applying sparse regularization on the disturbance $\boldsymbol{\delta}$ instead of the decoupled mask $\mathbf{m}$ cannot lead to a real sparse solution. 

\paragraph{Effects of quantization.}
Then we turn to the quantization operator. 
To show the effectiveness of our proposed operator, we compare it with training without quantization ($p$ = 0). Results show that without quantization in training, the attack can still achieve a competitive sparsity but the transferability drops. We further compare with straight through estimator (STE)~\cite{bengio2013estimating}. STE is a popular technique in binary neural networks~\cite{hubara2016binarized} to address the gradient problem occurring when training deep networks binarized by function. It chooses the identity function to approximate the derivative of the sign function. We find that the proposed method is better than STE in both sparsity and transfer rate.

\paragraph{Effects of losses.} We study the effect of sparse loss and quantization loss. Without sparse loss, the attack degrades to a dense attack with a 100$\%$ sparsity. Without quantization loss, the sparsity raises and meanwhile the transferability decreases, which proves our hypothesis that the information loss in quantization will lead to test error.

\paragraph{More data boosts transferability.} Our generator is trained with a large amount of data because we assume that more data can help alleviate overfitting. Here we empirically prove this assumption by comparing our method with `ad-hoc'. The `ad-hoc' setting means that, for each test image $\mathbf{x}$, we utilize the architecture in Figure~\ref{fig:1} to optimize a generator, which learns a mapping between $\mathbf{x}$ and its corresponding adversarial image $\mathbf{x}_{adv}$. The output of this generator, i.e., the generated adversarial image $\mathbf{x}_{adv}$, is then used for evaluation. The `ad-hoc' optimizes a generator only with a single image, while our method train a generator with a number of natural images. The result in Tabel~\ref{table:5} shows that `ad-hoc' achieves a higher white-box success rate but a lower transfer rate. This demonstrates adversarial examples crafted by `ad-hoc' are easier to overfit the white-box model. By training with more data, our method alleviates the overfitting and promotes the transferability.

\begin{table}
  \caption{Ablation study of the proposed framework. Results are sparsity($\%$) and fooling rate($\%$) on different models (fooling rate is not studied if sparsity is not satisfactory). `*' means white-box setting.}
  \label{table:5}
  \centering
  \begin{tabular}{lccccc}
    \toprule
    Method & Sparsity & IncV3 & Res50 & VGG16 & Dense161           \\
    \midrule
    The proposed  & { 0.46}  & { 61.24*} & { 63.76}  & { 85.94}  & { 46.22}  \\
    \midrule
    w/o decoupling & 82.88 &-&-&-&-\\
    $p$ = 0 & 0.47 & 19.92* & 29.06 & 50.88 & 21.44\\
    $q$ = STE~\cite{bengio2013estimating} & 3.26 & 39.30* & 50.06&71.36&38.16\\
    w/o sparse loss & 100.00 &-&-&-&-\\
    w/o quantization loss & 0.93 & 53.04* & 51.94 & 75.98 & 40.92\\
    \midrule
    ad-hoc & 0.47 &87.07*& 43.97 & 65.52 &23.28\\
    
    \bottomrule
  \end{tabular}
\end{table}


\section{Conclusion}
\label{sec:5}
In this paper, we propose a generator-based sparse adversarial attack framework. Under the same sparsity setting, it can achieve stronger transferability than existing state-of-the-art methods with a faster inference speed. Empirically, we observe that our approach leads to state-of-the-art results when generating attacks on the large scale ImageNet. Our work sheds light on the existence of transferable $\ell_0$ based sparse adversarial examples and illustrates state-of-the-art white-box sparse attack methods tend to find adversarial examples which have the least number of modified pixels but do not transfer. Both types of sparse adversarial attack are equally important to analyse the vulnerability of DNNs and evaluate security risks such as creating transferable adversarial patches in the physical world to deceive autonomous cars.

{
\small
\bibliographystyle{plain}
\bibliography{main.bib}
}

\end{document}